\pdfoutput=1

\documentclass[11pt]{article}

\PassOptionsToPackage{table}{xcolor}  
\usepackage{colortbl}                 

\usepackage[final]{acl}

\usepackage{times}
\usepackage{latexsym}
\usepackage{enumitem}
\usepackage[T1]{fontenc}

\usepackage[utf8]{inputenc}

\usepackage{microtype}

\usepackage{inconsolata}

\usepackage{graphicx}

\usepackage{marvosym}
\usepackage{multirow}

\usepackage{algorithm}
\usepackage[noend]{algpseudocode}

\algtext*{EndIf} 
\algtext*{EndFor} 
\algtext*{EndWhile} 

\usepackage{makecell}
\usepackage[most]{tcolorbox} 
\usepackage{array}
\usepackage{tabularx} 
\usepackage{caption} 

\usepackage{booktabs}
\usepackage{lineno}     
\usepackage{pifont}

\title{ZoomEye: Enhancing Multimodal LLMs with Human-Like Zooming Capabilities through Tree-Based Image Exploration}

\author{%
    \textbf{Haozhan Shen$^{1}$} \quad
    \textbf{Kangjia Zhao$^{1}$} \quad
    \textbf{Tiancheng Zhao$^{2,3}$\textsuperscript{\Letter}} \quad
    \textbf{Ruochen Xu$^{2}$} \\[1pt]
    \textbf{Zilun Zhang$^{1}$} \quad
    \textbf{Mingwei Zhu$^{1}$} \quad
    \textbf{Jianwei Yin$^{1}$}\\[3pt]
    $^1$ Zhejiang University \quad
    $^2$ Om AI Research \quad
    $^3$ Binjiang Institute of Zhejiang University \\
    \small{
       \Letter\ Correspondence: {tianchez@zju-bj.com}
     }
    \vspace{-3mm}
}

\begin{document}
\maketitle

\begin{abstract}
Multimodal Large Language Models (MLLMs) have demonstrated impressive capabilities in vision-language understanding. Recently, with the integration of test-time scaling techniques, these models have also shown strong potential in visual reasoning. However, most existing reasoning approaches remain text-level in nature: MLLMs are prompted to explore various combinations of textual tokens via their underlying language model, while the visual input remains fixed throughout the reasoning process. This paradigm limits the model’s ability to fully exploit rich visual information, particularly when dealing with images containing numerous fine-grained elements. In such cases, vision-level reasoning becomes crucial---where models dynamically zoom into specific regions of the image to gather detailed visual cues necessary for accurate decision-making. In this paper, we propose Zoom Eye, a training-free, model-agnostic tree search algorithm tailored for vision-level reasoning. Zoom Eye treats an image as a hierarchical tree structure, where each child node represents a zoomed-in sub-region of its parent, and the root corresponds to the full image. The algorithm enables MLLMs to simulate human-like zooming behavior by navigating from root to leaf nodes in search of task-relevant visual evidence. We experiment on a series of elaborate high-resolution benchmarks and the results demonstrate that Zoom Eye not only consistently improves the performance of a series of MLLMs with large margin~(e.g., InternVL2.5-8B increases by 15.71\% and 17.69\% on HR-Bench) but also enables small 3-8B MLLMs to outperform strong large models such as GPT-4o. Our code is available at \href{https://github.com/om-ai-lab/ZoomEye}{https://github.com/om-ai-lab/ZoomEye}.
\end{abstract}

\section{Introduction}
\label{sec:intro}

\begin{figure}[ht!]
    \centering
    \includegraphics[width=7.5cm]{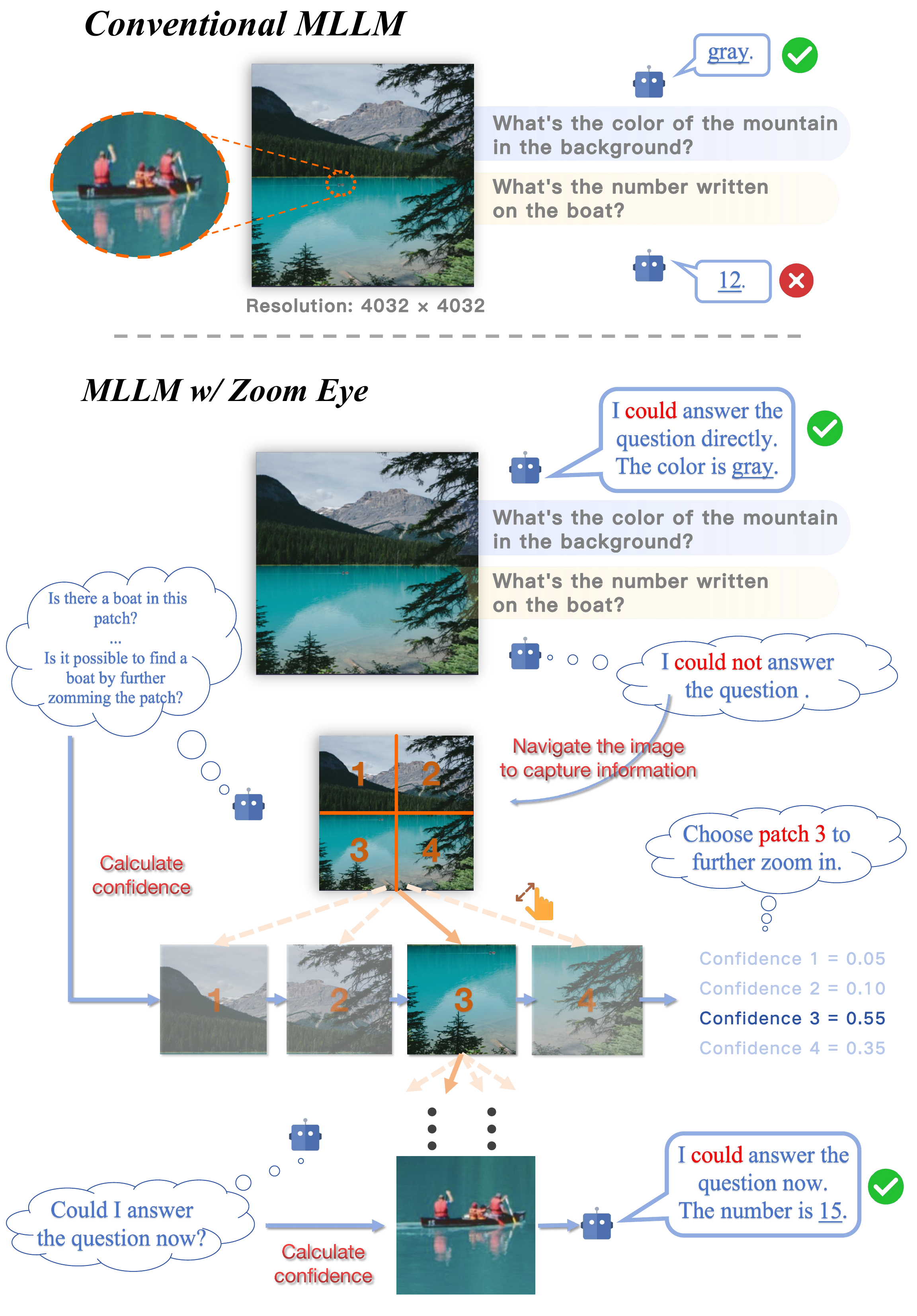}
    \caption{\textbf{Top}: When dealing with a high-resolution image, MLLMs effectively perceive the dominant objects but often fail to recognize finer details, highlighting the need for vision-level reasoning. \textbf{Bottom}: Applied with Zoom Eye, MLLMs could perform vision-level reasoning, allowed to explore the image details until they can answer the question.}
    \label{fig:introduce}
    \vspace{-3mm}
\end{figure}

By integrating powerful language models~\cite{touvron2023llama2, yang2024qwen2} with visual encoders~\cite{radford2021learning, EVA-CLIP, zhai2023sigmoid}, Multimodal large language models~(MLLMs) are able to jointly process textual and visual inputs, achieving impressive performance in vision-language understanding~\cite{zhao2024omchat,bai2023qwenvl,chen2024internvl,li2024llava}. Recently, drawing on test-time scaling techniques that enhance reasoning abilities in LLMs, such as OpenAI-o1~\cite{jaech2024openai} and DeepSeek-R1~\cite{guo2025deepseek}, a series of literature tries to investigate these reasoning techniques in MLLMs to further improve the visual reasoning capabilities~\cite{xu2024llavacot,dong2024insight,yao2024mulberry,shen2025vlm,meng2025mm}

However, these methods predominantly operate at the textual level, leveraging the generative capacity of the underlying language model without modifying the perception of the image itself. That is, the visual input remains static throughout the reasoning process, restricting the model’s ability to process fine-grained visual content, especially on an elements-rich high-resolution image. As illustrated in the top of Figure \ref{fig:introduce}, for the same image, the MLLM accurately recognizes the dominant object whereas it struggles to perceive the detailed one. This gap highlights the need for vision-level reasoning, where the model actively interacts with the image by zooming in and out to selectively attend to informative regions, as demonstrated in the bottom of Figure \ref{fig:introduce}, much like how humans visually process complex scenes. A similar vision-level zooming mechanism has been adopted in the closed-source OpenAI-o3~\cite{openai2025systemcard}. In contrast, our goal is to develop an open-source vision-level reasoning method, making this capability accessible to the broader research community.

\begin{figure}[t]
    \centering
    \includegraphics[width=7.7cm]{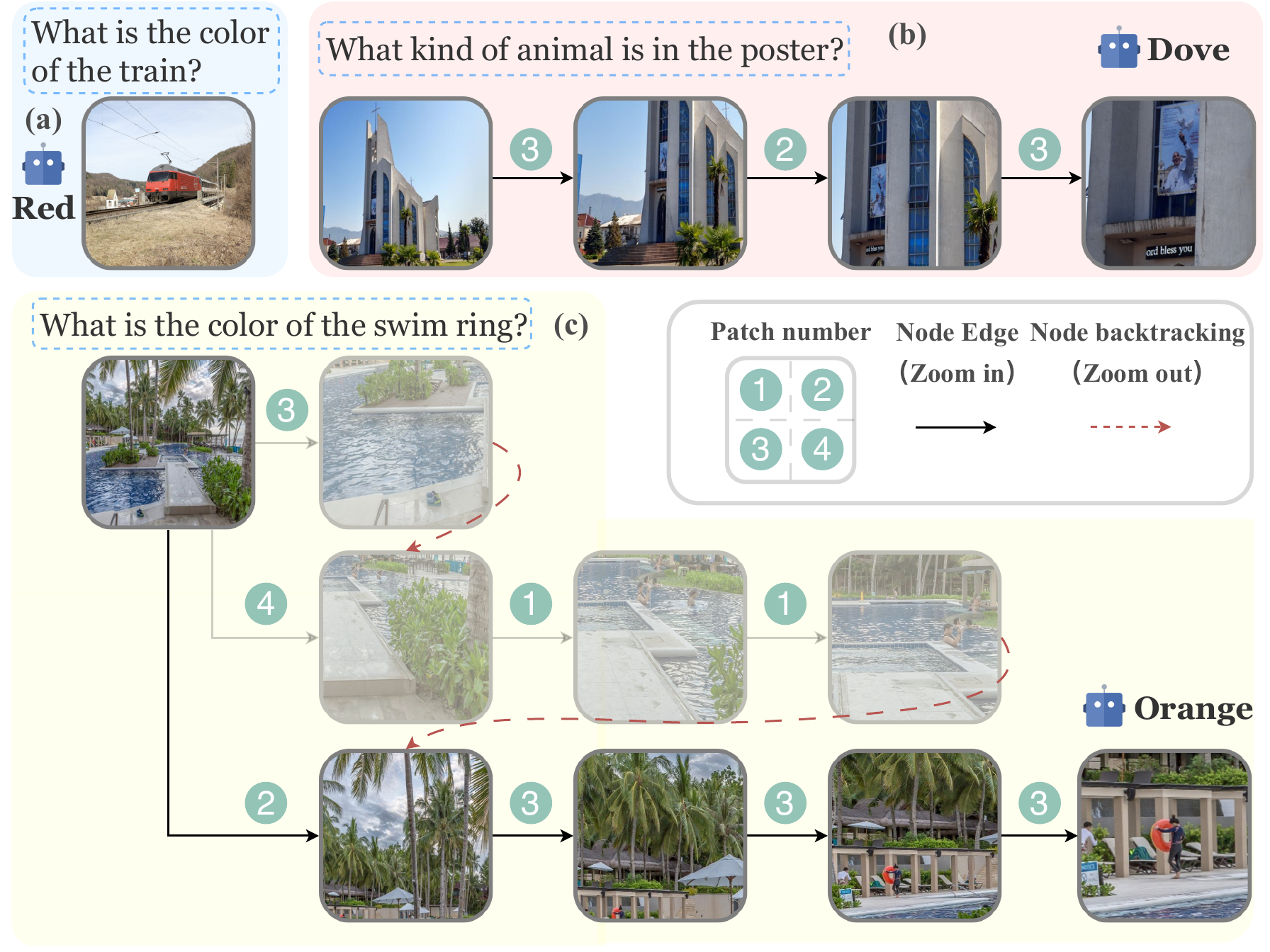}
     \caption{Zoom Eye enables MLLMs to \textbf{(a)}~answer the question directly when the visual information is adequate, \textbf{(b)}~zoom in gradually for a closer examination, and \textbf{(c)}~zoom out to the previous view and explore other regions if the desired information is not initially found.}
    \vspace{-2mm}
    \label{fig:example}
    \vspace{-2mm}
\end{figure}

When viewing a high-resolution image, humans typically start with a global scan, then gradually zoom into areas of interest for closer inspection (Figure~\ref{fig:example}(b)). If the desired information is not found, they zoom out and explore alternative regions~(as shown in Figure \ref{fig:example}~(c)). Inspired by this, structuring an image as a tree is highly logical for simulating similar actions in an MLLM: the root denotes the full image, each child node corresponds to a zoomed-in sub-region of its parent, and deeper nodes indicate higher zoom levels. This hierarchical representation, combined with a search algorithm, allows models to (1)\;explore fine-grained regions (\textit{node lookahead}) and (2)\;return to the previous view to inspect other regions (\textit{node backtracking}). Similar tree-based search strategies have shown strong performance in text-based LLM reasoning\cite{yao2024tree, hao2023reasoning, feng2023alphazero, zhu2023solving}.


In this paper, we propose Zoom Eye, a tree search algorithm for vision-level reasoning, which navigates MLLMs in the dense image context by the hierarchical and visual nature of images~(\textbf{contribution \#1}).  This method simulates the actions of zooming in and out to inspect image details and seek out crucial information. Given a question, the adopted MLLM first identifies the pertinent objects. We then introduce
two types of \textit{confidence values} by prompting the MLLM to recognize the presence of these relevant objects. These \textit{confidence values} are used to prioritize each candidate node during the tree search, determining the sequence of node selection.   The search concludes based on a stopping criterion  when the MLLM can confidently answer the question. This process is illustrated in the bottom part of Figure \ref{fig:introduce}.  Finally, the MLLM formulates a final response based on the visual information gathered during the search.

We adapt Zoom Eye to a series of mainstream MLLMs, including Qwen2.5VL~\cite{bai2025qwen2}, LLaVA-v1.5~\cite{liu2024improved}, LLaVA-OneVision~\cite{li2024llava}, InternVL2.5~\cite{chen2024expanding}, and evaluate them on a suite of elaborate high-resolution visual understanding benchmarks. Equipped with Zoom Eye, all evaluated models achieve substantial performance improvements compared to the baseline~(\textbf{contribution \#2}). 

Additionally, our analysis also reveals certain deficiencies in visual understanding exhibited by these models, which we detail in §\ref{sec:shortcomings}~(\textbf{contribution \#3}). Addressing these limitations is part of our future work. More importantly, as discussed in §\ref{sec:test-time scaling}, we observe a vision-level test-time scaling phenomenon analogous to what has been observed in text-based LLMs: performance consistently improves with an increasing number of search steps. This finding suggests that vision-level reasoning benefits from deeper exploratory search and opens new avenues for scaling MLLM inference beyond static image perception~(\textbf{contribution \#4}).

\section{Preliminary}
In this section, we describe briefly the prevalently adopted image preprocessing methods and image-text input ways of MLLMs.


\noindent\textbf{Image preprocessing.} For a given image \textbf{I}, a \textbf{naive} processing style is to simply resize it to a preset fixed resolution and then feed it into a vision encoder to generate visual representations. These representations can be treated as visual tokens and subsequently passed to an LLM, enabling the model to perceive the visual content of \textbf{I}. Formally, this process can be expressed as: $ \textbf{v} = \mathcal{F}(R(\textbf{I})) = (\textit{$v_1,v_2,\dots,v_{L_v}$}) $, where $\mathcal{F}$ is the vision encoder, $R$ is the resize operation, and $L_v$ is the number of visual representations, which also corresponds to the number of visual tokens accepted by the LLM. Due to the constraints of the naive version’s fixed and limited resolution, another method, known as \textbf{AnyRes}, was introduced. It divides the original image into several equal-area blocks and imposes a maximum limit, $M$, on the number of divided blocks. The vision encoder then independently encodes each block and the overall image. Finally, all the encoded visual representations are integrated together. This allows flexible processing of various resolutions. Denoting $\textbf{I}^{(0)}$ as the whole image and  $\{\textbf{I}^{(1)},\dots,\textbf{I}^{(a)}\}\;(a\leq M)$ as the blocks, the AnyRes could be formulated as: $\textbf{v} = \mathcal{F}(A(\textbf{I})) = [\textbf{v}_0, \textbf{v}_1, \dots, \textbf{v}_a]$, where $A$ denotes the AnyRes operation and $\textbf{v}_i = \mathcal{F}(R(\textbf{I}^{(i)})) = (v_{(i,1)},v_{(i,2)},\dots,v_{(i,L_v)}),\ i=0,1,\dots,a$. 
 It is noteworthy that the naive method can be considered a special case of AnyRes when $a = 0$ .

 \noindent\textbf{Imga-Text joint input for MLLM.} Common MLLMs link a  vision encoder  to the pre-trained LLM via projection or alignment modules, allowing language generation through the autoregressive capabilities of their LLM base.  Specifically, given an image  \textbf{I}  and an input prompt \textbf{x}, \textbf{I} is first encoded into a set of visual representations as described in the previous sub-section. Subsequently, these visual representations, along with the text input,  are fed into the LLM base of the MLLM. Assuming the length of the output sequence and text input are $L_y$ and $L_x$ respectively, the probability for a MLLM $\Phi_{\theta}$ to generate an output \textbf{y} = (\textit{$y_1,y_2,\dots,y_{L_y}$}) conditioned on the visual input $\mathcal{F}(\cdot(\textbf{I})) = (v_{(0,1)},\dots,v_{(a,L_v)})$ and the text input \textbf{x} = (\textit{$x_1,x_2,\dots,x_{L_x}$}) is: $\Phi_{\theta}(\textbf{y}|\mathcal{F}(\cdot(\textbf{I})),\textbf{x}) = \prod_{i=1}^{L_y} \Phi_{\theta}(y_i |v_{(0,1):(a,L_v)},x_{1:L_x}, y_{1:i-1})$, where $\mathcal{F}(\cdot)$ could represent $\mathcal{F}(R)$ as naive resize or $\mathcal{F}(A)$ as AnyRes. 
\section{Methodology}


In this section, we introduce the Zoom Eye algorithm. Firstly, we brief the general tree search algorithm. Subsequently, we elaborate on our implementation by initializing the components of the tree search algorithms in detail.
\subsection{Abstraction of Tree Search}
\textbf{Tree node.} Typically, a node in the tree structure comprises the following attributes:(1)~\textbf{id}: The unique identifier of the node.
(2)~\textbf{depth}: Represents the level of the node within the tree. 
(3)~\textbf{value}: Used to store numeric or textual data in the node.
(4)~\textbf{children}: A list of references to the node's children nodes,
  which facilitates traversal  of the tree structure.
(5)~\textbf{Other custom attributes}

\noindent\textbf{Tree search.} The abstraction of the tree search algorithm could be modeled as a tuple ($T,Q,\mathcal{R},\mathcal{S}$), where $T$ is the tree structure consisting of a set of nodes, $Q$ is a container that holds all the nodes that might be accessed in the next search step, $\mathcal{R}$ is a ranking function used to select the highest priority node based on the used search algorithm, and $\mathcal{S}$ represents the stopping criterion. The abstract search process is shown in Algorithm \ref{algorithm:abstract}.

\begin{algorithm}
\floatname{algorithm}{Alg.}
\caption{Abstraction of Tree Search Algorithm}
\small
\begin{algorithmic}[1] 
\Require{$T,Q,\mathcal{R},\mathcal{S}$} 
    \State Initialize $Q$ as the empty queue \{\}
    \State $Q$.\texttt{append($T$.root)}
    \While{$Q$ is not empty}
        \State $n_t \gets Q.\texttt{pop()}$
        \If{$\mathcal{S}(n_t)==\text{True}$}
            \State \textbf{break}
        \EndIf
        \State $s \gets n_t.\texttt{children.size}$
        \For{$j=1,\dots,s$}
            \State $Q$.\texttt{append($n_t$.children[j])}
        \EndFor
        \State $Q$.\texttt{sort($\mathcal{R}$)}
    \EndWhile
    
\end{algorithmic}\label{algorithm:abstract}
\end{algorithm}

Consider the example of a DFS search for a node with a \texttt{value} of 5 in the tree, in this case, $\mathcal{R}$ is a function that sorts the nodes in $Q$ in descending order of \texttt{depth}, and in ascending order of \texttt{id} when depths are equal. Meanwhile, $\mathcal{S}$ is a function checking if a node’s \texttt{value} equals 5.

A specific implementation of Zoom Eye search involves three key questions:
1. How to formulate the image as a tree $T$\;(§\ref{sec:tree}).
2. How to set the ranking function $\mathcal{R}$\;(§\ref{sec:confidence}).
3. How to determine the stopping criterion $\mathcal{S}$\;(§\ref{sec:stop}).
Finally, we provide a description of the overall algorithm in §\ref{sec:algorithm}.

\begin{figure}[ht!]
    \centering
    \includegraphics[width=7cm]{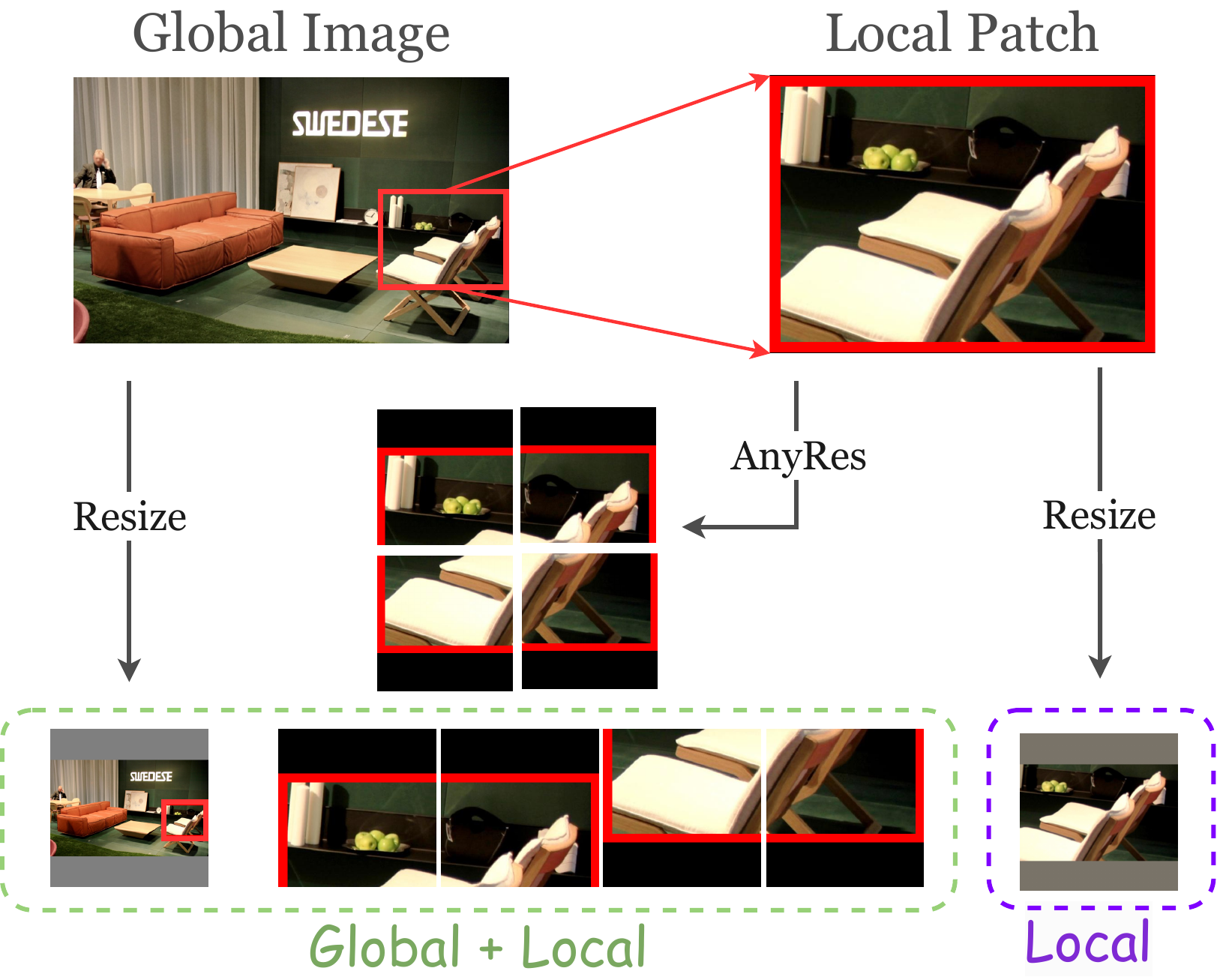}
    \caption{Two image input methods for MLLMs with distinct image processing.}
    \label{fig:image input}
\end{figure}

\subsection{Tree Representation for Image}\label{sec:tree}
We model the overall image as a tree $T$. A specific \textit{node}, denoted as $n_t$, represents an image patch view $\{\textbf{I}, \textbf{b}_t\}$, where \textbf{I} is the image and $\textbf{b}_t = (x_{1,t},y_{1,t},x_{2,t},y_{2,t})$ is the normalized bounding box coordinates. If the size of $n_t$’s image patch exceeds the predefined resolution by the image encoder, it can be further divided into four equal-sized sub-patches, serving as its children with size 4. Nodes are recursively divided until they meet the resolution limit. At the start of the search, the root node $T.\texttt{root} = \{\textbf{I}, (0, 0, 1, 1)\}$ representing the overall image is visited.

However, due to the detailed nature of high-resolution images and information loss from downsampling to the vision encoder’s fixed resolution, MLLMs frequently struggle to accurately capture key parts of an image initially. Consequently, MLLMs should be be allowed to  continuously scan and zoom into the current view (i.e., explore deeper nodes) for more focused information. In our implementation, we consider two image input methods to enable MLLMs to perceive the local patch represented by  $n_t$:\;(1)\;Local Input: only the local patch is provided, suitable for earlier single-image input MLLMs with naive image preprocessing method~\cite{li2023blip,liu2024visual,liu2024improved}.\;(2)\;Global+Local Input: both the global image and local patch are input, ideal for advanced MLLMs using AnyRes preprocessing method~\cite{liu2024llavanext,li2024llava,chen2024internvl}. In this case, we use the visual prompt with a red rectangle to emphasize the local focus, applying naive processing to the global image and AnyRes to the local patch, as shown in Figure \ref{fig:image input}. Denoting $\mathcal{V}(n_t)$ as the final image input, we have:

\vspace{-3mm}
{\footnotesize
\begin{equation}
\mathcal{V}(n_t) = 
\begin{cases}
    \textstyle [\mathcal{F}(R(\textbf{I}.\text{crop}(\textbf{b}_t))] & \text{Local} \\
    \textstyle [\mathcal{F}(R(\textbf{I})), \mathcal{F}(A(\textbf{I}.\text{crop}(\textbf{b}_t))] & \text{Global+Local}
\end{cases}
\label{eq vt}
\end{equation}
}

\begin{algorithm}
\floatname{algorithm}{Alg.}
\caption{Ranking Function\;\&\;Stopping Criterion}
\small
\begin{algorithmic}[1] 
\Require{$\Phi_{\theta}, \mathcal{W}, \{\text{p}_e, \text{p}_l, \text{p}_a\}, \tau, o, q_s$}
\Function{$\mathcal{R}$}{$n_1$, $n_2$}\Comment{Ranking Function}
    \State \Return \Call{get priority}{$n_1$} $>$ \Call{get priority}{$n_2$}
\EndFunction
\State
\Function{$\mathcal{S}$}{$n_t$}\Comment{Stopping Criterion}
    \State $c_a \gets \Call{Logits Ratio}{n_t, \text{p}_a(q_s)}$
    \State \Return $c_a \ge \tau$
\EndFunction
\State
\Function{get priority}{$n_t$}
    \If{$n_t$.\texttt{priority} is None}
        \State $c_e \gets \Call{Logits Ratio}{n_t, \text{p}_e(o)}$
        \State $c_l \gets \Call{Logits Ratio}{n_t, \text{p}_l(o)}$
        \State $\alpha \gets \mathcal{W}(n_t.\texttt{depth})$ \Comment{weighted factor}
        \State $n_t.\texttt{priority} \gets \alpha\cdot c_l + (1-\alpha)\cdot c_e$
    \EndIf
    \State \Return{$n_t$.\texttt{priority}}
\EndFunction
\State
\Function{Logits Ratio}{$n_t$, \textbf{x}}
    \State $z_1 \gets \Phi_{\theta}(y=\texttt{Yes}\mid\mathcal{V}(n_t),\;\textbf{x})$
    \State $z_2 \gets \Phi_{\theta}(y=\texttt{No}\mid\mathcal{V}(n_t),\;\textbf{x})$
    \State $z \gets (softmax(z_1, z_2)[0]-0.5) \times 2)$ 
    \State \Return{$z$} \Comment{$z \in (-1,1)$}
\EndFunction
\end{algorithmic}\label{algorithm:rank}
\end{algorithm}
\vspace{-5mm}

\subsection{Ranking Function} \label{sec:confidence}
As shown in Algorithm \ref{algorithm:abstract}, $\mathcal{R}$ is used to rank the nodes with the priority value to determine which one to visit  in the next step. A well-defined $\mathcal{R}$ strategically steers the search process. In Zoom Eye, we adopt the MLLM to calculate the priority value and use $\mathcal{R}$ to sort nodes by the value. Specifically, let $o$ denote the visual cue that is crucial for answering the question, a MLLM should have the following capabilities: (1)\;It could perceive whether $o$ exists within the visible view; (2)\;If $o$ occupies a small area and is not clearly visible, it can leverage the common sense knowledge to infer whether $o$ might be discerned through further zooming. Thus, we query the MLLM with two prompts $\text{p}_e(o)$ and $\text{p}_l(o)$\;(e.g., ``\texttt{Is there a $o$ in the sub-patch?}", ``\texttt{Is it possible to find a $o$ by further zooming the sub-patch?}") to trigger these two capabilities, and use the ratio of the next-word probability of the token ``\texttt{Yes}" and ``\texttt{No}" as priority values. We refer to these two values as \textit{existing confidence} and \textit{latent confidence}, denoted as $c_e$ and $c_l$.

The overall priority value for a node is the weighted sum of  $c_e$ and $c_l$. We introduce a weight function $\mathcal{W}(d)$  that is related to a node's \texttt{depth}. When the depth is shallow, indicating minimal zoom and the MLLM might not clearly perceive the cue, assign more weight $c_l$. As depth increases, shift more weight to $c_e$. Finally, ranking function $\mathcal{R}$ is introduced to rank nodes by the overall priority value, as shown in Algorithm \ref{algorithm:rank}.

\subsection{Stopping Criterion}\label{sec:stop}
Zoom Eye exits the search process when the MLLM provides feedback that the current view is sufficient to answer the provided question, denoted as $q_s$. Specifically, we query the MLLM with a prompt $\text{p}_a(q_s)$\;(e.g., ``\texttt{Could you answer $q_s$ now?}") and use the same method as described in §\ref{sec:confidence} to quantify the positive feedback. We refer to it as \textit{answering confidence}, denoted as $c_a$. When $c_a$ exceeds a predefined threshold $\tau$, the search terminates. The implementation of $\mathcal{S}$ is shown in Algorithm \ref{algorithm:rank}.

\subsection{Overall Search Algorithm}\label{sec:algorithm}
With the above notations in place, we now describe how Zoom Eye works for a given image-question pair (\textbf{I}, $q$). The complete algorithm workflow is shown in Appendix~\ref{sec:complete}.

\noindent\textbf{Generating visual cues to guide the search.} Before search, the MLLM has to predefine the visual cues essential for addressing $q$, enabling a targeted and guided search based on these cues. We utilize the in-context capability from the LLM base of the MLLM, using a sequence of contextual examples as prefixes to generate visual cues. Ultimately, the MLLM produces $k$ visual cues \(\{o_1, \dots, o_k\}\) pertinent to $q$. Each $o_i$ ($i \in \{1,\dots,k\}$) can be categorized into two types: (type\,1) those requiring a search for a single instance, and (type\,2) those requiring identification of all instances in the image. 

\begin{table}[h!]
\footnotesize
  \begin{center}
  \resizebox{\hsize}{!}{
  \setlength\tabcolsep{4pt}
    \begin{tabular}{l c c c c c c} 
    \toprule
                 &    & \textbf{Question}       &  & \textbf{Visual cues} & &\textbf{Type}\\
    \midrule
              1     & \vline  & {What is the color of the dog?}     & \vline& \textit{dog} & \vline & type\,1  \\
     \midrule
              \multirow{2}{*}{2}     & \vline  & \multirow{2}{*}{\makecell{{What is the relative position}\\ {of the dog to the cat?}}}   & \vline  & \multirow{2}{*}{\makecell{\textit{dog},\\ 
 \textit{cat}}}  & \vline & \multirow{2}{*}{\makecell{type\,1,\\type\,1}} \\
                 & \vline  &     & \vline & & \vline & \\
    \midrule
              3  & \vline  &  {How many dogs in the image?}   & \vline & \textit{all dogs} &\vline & type\,2 \\

    \bottomrule
    \end{tabular}
    }
  \end{center}
  \vspace{-2mm}
  \caption{Examples of visual cues and their types.}
  \label{table:cues type}
  \vspace{-2mm}
\end{table}

\noindent\textbf{Searching for cues.} For each cue $o_i$ ($i \in \{1,\dots,k\}$), Zoom Eye explores the image tree to capture pertinent visual information. When searching for type 1 cues, the search is guided with $\mathcal{R}$ and concludes as soon as it meets $\mathcal{S}$, then the current node is recorded in a list $L$ . For a single type\,1 clue, as shown in line 1 of Table \ref{table:cues type}, the applied $q_s$ for$\mathcal{S}$  is the input question $q$. If multiple type\,1 clues are generated as in line 2 of Table \ref{table:cues type}, we introduce a decomposed question template $\text{p}_{dq}(o_i)$ such as ``\texttt{what is the location of the $\{o_i\}$?}" specific to each cue. In this case, the applied $q_s$ of $o_i$ is $\text{p}_{dq}(o_i)$. If a type\,2 cue is generated, as shown in line 3 of Table \ref{table:cues type}, $\mathcal{S}$ is not applied, and we search the whole tree to add all nodes with sufficient \textit{existing confidence} to  $L$.

\noindent\textbf{Answering the question using the searched cues.}  Given the searched nodes $ L = \{n^*_1,\dots,n^*_K\}$ , the MLLM formulates a response to the input question $q$ by synthesizing information of these nodes. Denoting $\textbf{b}^*_i = (x^*_{1,i},y^*_{1,i},x^*_{2,i},y^*_{2,i})$ as the bounding-box of $n^*_i$ ($i \in \{1,\dots,K\}$), we union the bounding-box coordinates of all nodes in $L$ to create a union bounding-box $\textbf{b}^* = ( \min_i x^*_{1,i}, \min_i y^*_{1,i}, \max_i x^*_{2,i}, \max_i y^*_{2,i} )$. For the two distinct image input methods, we apply  Eq. \ref{eq vt} to feed the focused region $\textbf{b}^*$ along with $q$ into models and derive the final response.

\section{Experiments}

\subsection{Implementation Details}\label{sec:Implementation}
\noindent\textbf{Local input.} We select LLaVA-v1.5-7B~\cite{liu2024improved} as the base MLLM, with the naive image processing. We set $\tau$ at 0.8 and define $\mathcal{W}$ as $ \frac{1-b}{D^2}\times d^2 + b $, where $D$ denotes the depth of the image tree, $d$ is the depth of the visited node during the search, and $b$ is a bias value, set here at 0.2. 


\noindent\textbf{Global\,+\,Local input.} We select Qwen2.5VL-3B~\cite{bai2025qwen2}, LLaVA-ov(oneVision)-7B~\cite{li2024llava}, and InternVL2.5-8B~\cite{chen2024expanding} as our MLLMs, with the AnyRes image processing. For LLaVA-ov and InternVL, we define the maximum AnyRes block as 12, and for QwenVL, we set the max pixels as 12,~845,~056. We set $\tau$ at 0.6 and define $\mathcal{W}$ similarly to the above, except with $b$ of 0.6.

\noindent\textbf{For both input implementation}, we set the maximum search depth at 2 when searching for type\,2 cues to save costs. Additionally, the decomposed question template $\text{p}_{dq}(o_i)$ is assigned as ``\texttt{What is the appearance of the $\{o_i\}$?}". More details are described in Appendix~\ref{sec:appendix Implementation}.

\definecolor{darkgreen}{RGB}{0, 150, 0}
\definecolor{darkred}{RGB}{150, 0, 0}

\begin{table*}[h!]
\scriptsize
  \begin{center}
    \begin{tabular}{l c c c c c c c c c c c} 
    \toprule
       & \multicolumn{3}{c}{$V^*$ Bench} & & \multicolumn{3}{c}{HR-Bench 4K} & & \multicolumn{3}{c}{HR-Bench 8K}\\
      \textbf{Model} & \textbf{Attr} & \textbf{Spatial} & \textbf{Overall} & & \textbf{FSP} & \textbf{FCP} & 
      \textbf{Overall} & & \textbf{FSP} & \textbf{FCP} & \textbf{Overall}\\
    \midrule
        \textbf{Open-source MLLMs} & &  & & \vline &  &  &  & \vline &  & & \\ 
        minigptv2-7B~\cite{chen2023minigpt} &- & - & - & \vline & 25.75 & 25.25 & 25.50 & \vline & 26.0 & 26.25 & 26.13 \\
        LLaVA-v1.6-7B~\cite{liu2024llavanext} & 60.87 & 63.16 & 61.78 & \vline & 49.0 & 46.75 & 47.88 & \vline & 37.25 & 44.25 & 40.75 \\
        LLaVA-v1.6-13B~\cite{liu2024llavanext} &60.0 & 64.47 & 61.78 & \vline & 49.75 & 41.25 & 45.50 & \vline & 38.0 & 38.25 & 38.13 \\
        Yi-VL-34B~\cite{ai2024yi} &- & - & - & \vline & 46.0 & 42.75 & 44.38 & \vline & 39.50 & 38.50 & 39.0 \\
        LLaVA-HR-X-7B~\cite{luo2024feast} &51.30 & 64.47 & 56.54 & \vline & 57.75 & 46.25 & 52.0 & \vline & 42.0 & 41.25 & 41.63 \\
    \midrule
        \textbf{Closed-source MLLMs} & &  & & \vline &  &  &  & \vline &  & & \\       QWen-VL-max~\cite{bai2023qwenvl} &- & - & - & \vline & 65.0 & 52.0 & 58.50 & \vline & 54.0 & 51.0 & 52.50 \\
        GPT4o~\cite{achiam2023gpt}  &- & - &66.0 & \vline & 70.0 & 48.0 & 59.0  & \vline & 62.0  & 49.0 & 55.5 \\
        
    \midrule
      \textbf{Baseline and Local Input Zoom Eye} & &  & & \vline &  &  &  & \vline &  & \\
      LLaVA-v1.5-7B~\cite{liu2024improved}            &43.47 & 56.57 & 48.68& \vline & 38.5 & 33.75 & 36.13 & \vline & 33.0   & 31.25 & 32.13\\
      LLaVA-v1.5-7B w/ Zoom Eye & 83.45 & 82.89 & 83.25 & \vline & 67.75 & 38.75 & 53.25 & \vline & 65.50 & 36.0& 50.75 \\
      \rowcolor{gray!20}
      \quad \quad \quad \quad $\Delta$ & \textcolor{darkgreen}{+40.48}	&\textcolor{darkgreen}{+26.32}	& \textcolor{darkgreen}{+34.57}& \vline &  \textcolor{darkgreen}{+29.25} &	\textcolor{darkgreen}{+5.0 }&	\textcolor{darkgreen}{+17.12} & \vline &  \textcolor{darkgreen}{+32.50}	&\textcolor{darkgreen}{+4.75}	& \textcolor{darkgreen}{+18.62} \\
    \midrule
     \textbf{Baseline and Global+Local Input Zoom Eye} & &  & & \vline &  &  &  & \vline &  &  & \\
      
    
    Qwen2.5VL-3B~\cite{bai2025qwen2}         & 80.87&	71.05	&76.96 & \vline &  82.75&	49.0	& 65.88 & \vline &  80.5	& 45.25	& 62.88 \\
     Qwen2.5VL-3B  w/ Zoom Eye & 88.70&	\textbf{89.47} &	89.01 & \vline & 86.75 &	53.50&	70.13 & \vline &  84.75	&52.0 &	68.38 \\
     
    \rowcolor{gray!20}     
    \quad \quad \quad \quad $\Delta$ & \textcolor{darkgreen}{+7.83}	&\textcolor{darkgreen}{+18.42}	& \textcolor{darkgreen}{+12.05}& \vline &  \textcolor{darkgreen}{+4.0} &	\textcolor{darkgreen}{+4.50 }&	\textcolor{darkgreen}{+4.25} & \vline &  \textcolor{darkgreen}{+4.25}	&\textcolor{darkgreen}{+6.75}	& \textcolor{darkgreen}{+5.50} \\
     \arrayrulecolor{gray}\hline  
     \arrayrulecolor{black}       
     LLaVA-ov-7B~\cite{li2024llava}             & 75.65 & 75.0 & 75.39 & \vline & 72.0 & 54.0 & 63.0 & \vline & 67.25 & 52.25 & 59.75\\

     LLaVA-ov-7B w/ Zoom Eye  & \textbf{93.91} & 85.53 & \textbf{90.58} & \vline & 84.25 & 55.0 & 69.63 & \vline & 88.5 & 50.0 & 69.25 \\
     \rowcolor{gray!20}
      \quad \quad \quad \quad $\Delta$ & \textcolor{darkgreen}{+18.26}	&\textcolor{darkgreen}{+10.53}	& \textcolor{darkgreen}{+14.19}& \vline &  \textcolor{darkgreen}{+12.25} &	\textcolor{darkgreen}{+1.0 }&	\textcolor{darkgreen}{+6.63} & \vline &  \textcolor{darkgreen}{+21.25}	&\textcolor{darkred}{-2.25}	& \textcolor{darkgreen}{+10.0} \\
     \arrayrulecolor{gray}\hline  
     \arrayrulecolor{black}       
     InternVL2.5-8B~\cite{chen2024expanding}         & 67.83	&71.05	& 69.11& \vline &  75.75&	56.25&	66.0 & \vline &  61.5	&53.25	& 57.38 \\
     InternVL2.5-8B w/ Zoom Eye & 86.09&	82.89	& 84.82 & \vline & \textbf{88.75}	& \textbf{61.50} &	\textbf{75.13} & \vline &  \textbf{89.75}	&\textbf{57.5}&	\textbf{73.63} \\
     
    \rowcolor{gray!20}     
    \quad \quad \quad \quad $\Delta$ & \textcolor{darkgreen}{+18.26}	&\textcolor{darkgreen}{+11.84}	& \textcolor{darkgreen}{+15.71}& \vline &  \textcolor{darkgreen}{+13.0} &	\textcolor{darkgreen}{+5.25 }&	\textcolor{darkgreen}{+9.13} & \vline &  \textcolor{darkgreen}{+28.25}	&\textcolor{darkgreen}{+4.25}	& \textcolor{darkgreen}{+16.25} \\
    \bottomrule
    \end{tabular}
  \end{center}
  \vspace{-2mm}
  \caption{Results of different models on high-resolution benchmarks. FSP: Fine-grained Single-instance Perception; FCP: Finegrained Cross-instance Perception. More results are displayed in Table~\ref{table:appendix main results}.}
  \label{table:main results}
\end{table*}

\begin{table*}[h!]
\scriptsize
  \begin{center}
    \begin{tabular}{l c c c c c c c c c c c c} 
    \toprule
       & \multicolumn{5}{c}{\textbf{MO}} & \vline & \multicolumn{3}{c}{\textbf{AD}} & \vline& \multicolumn{2}{c}{\textbf{RS}} \\
       
       \textbf{Method} & Calculate & Intention & Property & Orientation  &Color$^\dagger$  & \vline  & Intention$^\dagger$ & Attention &  Motion$^\dagger$&   \vline&  Count & Position \\ 
      
    \midrule
        LLaVA-ov-7B &  36.33 &27.55 & 55.0 &  14.94& 34.19 & \vline  & 37.32 & 71.89 & 30.61 &  \vline& 32.95& 61.40 \\ 
        \quad w/ Zoom Eye & 38.67 & 38.78&  60.0& 14.62 & 47.09 & \vline  & 38.56 & 68.66 &  42.71&  \vline&35.56 &48.45  \\  
        \rowcolor{gray!20}
        \quad \quad \quad $\Delta$ & \textcolor{darkgreen}{+2.34} &\textcolor{darkgreen}{+11.23} & \textcolor{darkgreen}{+5.0} &  \textcolor{darkred}{-0.32}& \textcolor{darkgreen}{+12.90} & \vline  & \textcolor{darkgreen}{+1.24} & \textcolor{darkred}{-3.23} & \textcolor{darkgreen}{+12.10} &  \vline& \textcolor{darkgreen}{+2.61}& \textcolor{darkred}{-12.95} \\
    \bottomrule
    \end{tabular}
  \end{center}
  \vspace{-2mm}
  \caption{Performance comparison on MME-RealWorld benchmark. This benchmark comprises numerous sub-tasks, and we only list those that exhibit obvious performance changes of Zoom Eye against the baseline. MO\;(Monitoring), AD\;(Autonomous Driving), and RS\;(Remote Sensing) are data categories within this benchmark. $^\dagger$This result is an average derived from multiple similar sub-tasks (e.g., Color is the average of Vehicle Color and Person Color).}
  \label{table:realworld results}
  \vspace{-2mm}
\end{table*}

\subsection{Results on High-Resolution Benchmark}

\textbf{Evaluated benchmark.} We evaluate Zoom Eye on two meticulously curated high-resolution benchmarks. The first, $\textbf{V}^*$ \textbf{Bench}~\cite{wu2024v}, with an average resolution of 2246x1582, features sub-tasks in attribute recognition and spatial reasoning.  The second, \textbf{HR-Bench 8K}~\cite{wang2024divide} boasts average resolution of 7680, which consists of two sub-tasks: Fine-grained Single-instance Perception (FSP) and Fine-grained Cross-instance Perception (FCP). The 8K images are cropped around the objects in question to produce \textbf{HR-Bench 4K}. Both benchmarks are comprised of rich visual elements and required detailed perception to accurately respond. More results are displayed in Table~\ref{table:appendix main results}.

\noindent\textbf{Main results.} As shown in Table~\ref{table:main results}, all evaluated models exhibit significant performance gains after incorporating Zoom Eye, highlighting its model-agnostic applicability. For instance, LLaVA-ov-7B achieves performance improvements of 14.19\%, 6.63\%, and 10.00\% on $V^*$ Bench, HR-Bench 4K, and HR-Bench 8K, respectively.
In conjunction with the case studies presented in Figure~\ref{fig:case}, these results demonstrate that vision-level reasoning enables MLLMs to more effectively capture fine-grained and task-relevant visual information in complex scenes, thereby enhancing their overall visual understanding capabilities.

\subsection{Results on Real-World Benchmark}\label{sec:shortcomings}

\textbf{Evaluated benchmark.} We further evaluate Zoom Eye on MME-RealWorld~\cite{zhang2024mme}, a manually annotated benchmark tailored for real-world applications, featuring an average resolution of 2000$\times$1500. It includes 5 data categories and 43 sub-class tasks. Due to the page limit, we report on only 13 sub-tasks that show significant performance changes with Zoom Eye. These sub-tasks span 3 data categories, with similar types merged (e.g., Vehicle Color and Person Color into Color) to present average scores. Detailed results are provided in Appendix~\ref{sec:appendix MME-RealWorld}.

\noindent\textbf{Results.} As shown in Table \ref{table:realworld results}, Zoom Eye improves the performance of LLaVA-ov-7B on most sub-tasks, especially on MO/Intention\;(+11.23\%), MO/Color\;(+12.9\%), and AD/Motion\;(+12.1\%). However, we also notice that the model’s performance with Zoom Eye decline on some sub-tasks. We selecte one error example each from MO/Orientation and RS/Position and display them in Figure \ref{fig:case}. For MO/Orientation, the low direct response scores for LLaVA-ov, as seen in the Table \ref{table:realworld results}, along with error example in the figure, suggest a probable deficiency of orientation data during training, negatively impacting model performance in this aspect. For RS/Position, despite Zoom Eye locates the target, the final response was incorrect, suggesting the model struggles to link positional relationships between the full image and sub-images, resulting in a marked decline in performance on this sub-task. These error examples reveal the model’s deficiencies, by which we will guide the direction of improvements in the model’s capabilities in our future work.

\subsection{Ablation Studies}
\subsubsection{Vision-level test-time scaling}\label{sec:test-time scaling}

\begin{figure}[t]
  \centering
  \includegraphics[width=0.85\linewidth]{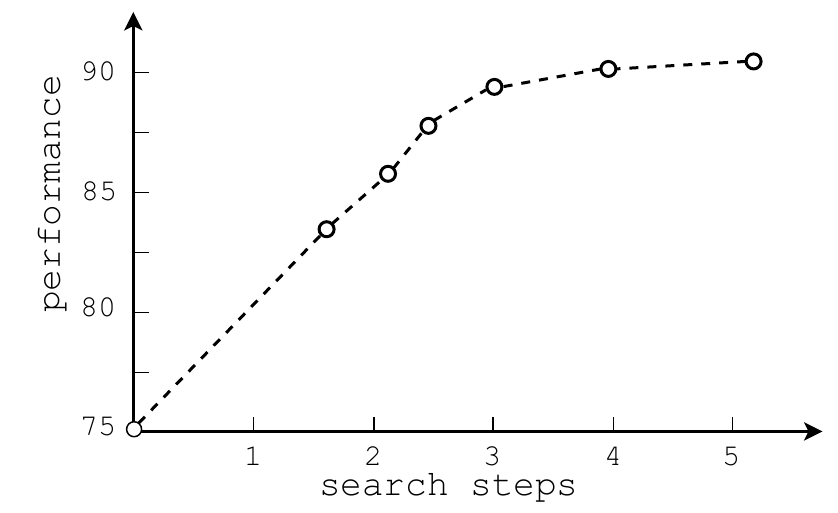}
   \caption{The relationship between the number of search steps and the performance of the MLLM. The experimental statistics are derived from LLaVA-ov-7B’s results on $V^*$ Bench.}
   \label{fig:infer}
\end{figure}

We progressively reduce the answering confidence threshold $\tau$ and analyze the relationship between the number of search steps and the performance of the MLLM, as illustrated in Figure \ref{fig:infer}.

From the figure, it can be seen that as the number of search steps increases, the model performance improves and eventually stabilizes.  This behavior is analogous to the test-time scaling in text-level reasoning, where the accuracy of the final answer improves with more CoT tokens being explored. This finding could be viewed as a form of \textbf{vision-level test-time scaling}, where exploring more detailed zoomed information instead of the static image could enhance the ability of MLLM to generate more accurate responses.

When deploying Zoom Eye in real-world scenarios, we can adjust the confidence threshold or the maximum number of search steps based on specific needs to achieve the best trade-off between performance and efficiency.

\begin{table}[t!]
\scriptsize
  \begin{center}

  \setlength\tabcolsep{8pt}
    \begin{tabular}{l c c c c} 
    \toprule
             \textit{Used MLLM} & & \textit{Zoom Successfully} &  & \textit{Performance} \\
    \midrule
        LLaVA-ov-7B   & \vline &  \ding{51} & \vline &  93.45 \\
        LLaVA-ov-7B   & \vline &  \ding{55} & \vline & 54.55 \\
    \bottomrule
    \end{tabular}

  \end{center}
  \caption{Comparison of MLLM performance conditioned on whether zoom is successful. A zoom is considered successful when the searched box covers at least 50\% of the target object. The experimental statistics are derived from $V^*$ Bench.}
  \label{table:zoom}
\end{table}

\subsubsection{Does the Zoom operation contribute to the improvement of the MLLM?}
By comparing the answer accuracy of MLLM when Zoom is successful versus when it fails, we investigate the contribution of the Zoom operation to the model. As shown in Table \ref{table:zoom}, the accuracy sees a remarkable improvement (from 54.55\% to 93.45\%) when Zoom is successfully applied. This substantial gain highlights the critical role of the Zoom operation. By effectively refining the model’s focus on relevant visual details, it contributes to more accurate and reliable responses, reinforcing its importance as a key mechanism for optimizing visual understanding.

\begin{table}[t!]
\footnotesize
\scriptsize
  \begin{center}

  \setlength\tabcolsep{3pt}
    \begin{tabular}{l c c c c c c} 
    \toprule
             \textbf{Model} & \textbf{Sub-region} &  & $V^*$ & HR-4K & HR-8K & Avg. Search \\
    \midrule
        LLaVA-ov-7B  & - & \vline & 75.39 &  63.00 & 59.75  & - \\
        \ \ w/ Zoom Eye  & 4 & \vline & 90.58  & 69.63	& 69.25	  & 8.20   \\
        \ \ w/ Zoom Eye  & 9 & \vline &  	93.19	& 69.75 & 	67.63	& 5.71     \\
        \ \ w/ Zoom Eye  & 16 & \vline &  92.15	& 70.38	 & 69.75	 &5.02      \\
    \bottomrule
    \end{tabular}

  \end{center}
  \caption{Comparison of MLLM performance conditioned on various number of the split sub-regions. Avg. Search means the number of average search steps in this setting.}
  \label{table:sub-region}
\end{table}

\subsubsection{Impact of the various number of the split sub-regions}
In this part, we conduct an ablation study to examine how the number of sub-regions split in the image tree affects the performance of Zoom Eye. The results are summarized in Table~\ref{table:sub-region}. We observe that, as the number of sub-regions increases, the performance of Zoom Eye improves slightly, while the number of search steps decreases. Overall, the results remain stable across different sub-region settings, suggesting that Zoom Eye is robust to variations in zooming granularity. These findings highlight the role of zooming granularity in the Zoom Eye algorithm.

\begin{figure*}[ht]
    \centering
    \includegraphics[width=16cm]{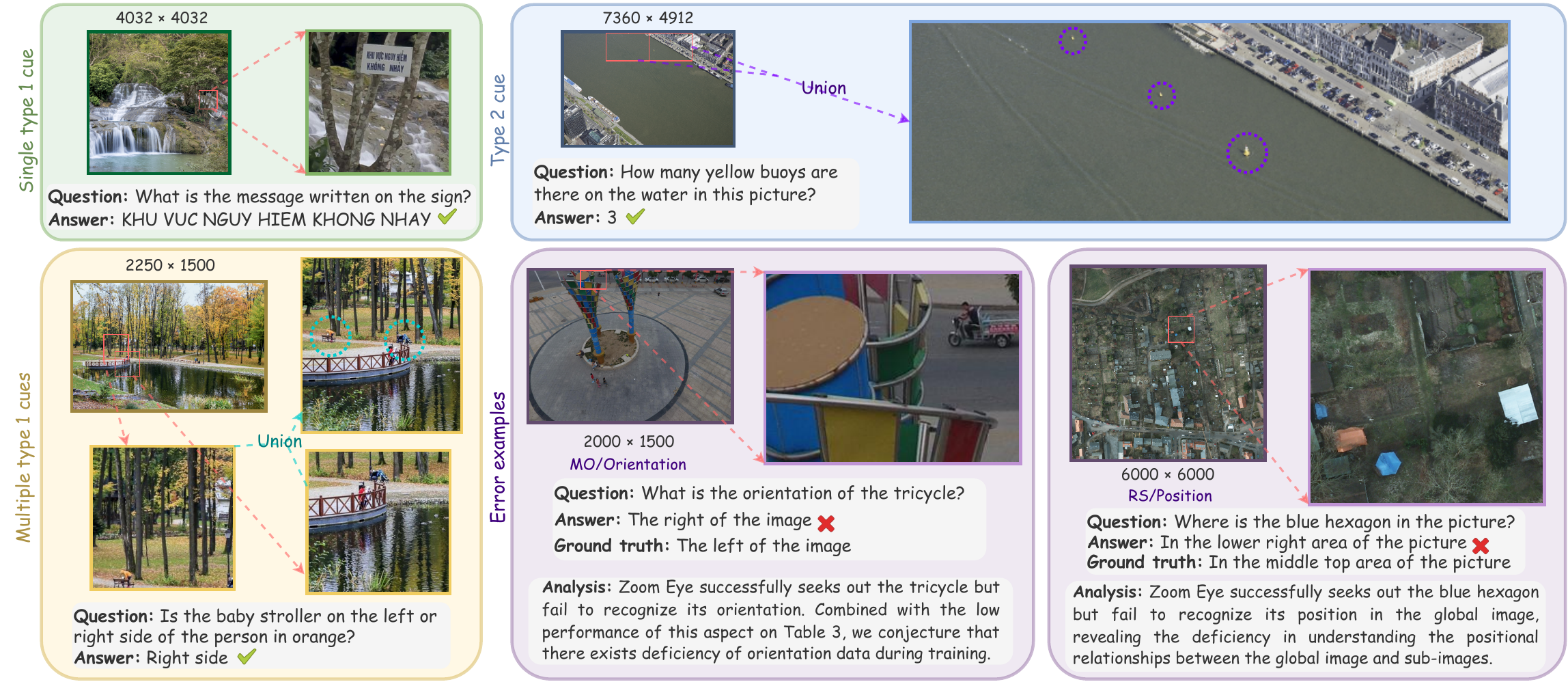}
     \caption{Examples of Zoom Eye. The resolution of the image is displayed. Red rectangles are patches searched by Zoom Eye.}
    \label{fig:case}
\end{figure*}

\subsection{Compared with Other HR Processing Methods}\label{sec:compare}

\begin{table*}[h!]
\scriptsize
  \begin{center}
    \begin{tabular}{l c c c c c c c c c c c c} 
    \toprule
       
      \textbf{Model} & & \textbf{Size} &  & \textbf{Method} & & \textbf{Training-free} & & \textbf{$V^*$ Bench}  & & \textbf{HR-Bench 4K} & & \textbf{HR-Bench 8K} \\
    \midrule
      \multirow{3}{*}{LLaVA-v1.5} & \vline & 7B & \vline & DC$^2$ & \vline & \ding{51} & \vline & 57.60 & & - &  & 39.50 \\
      & \vline & 7B & \vline & VisCrop & \vline & \ding{51} & \vline & 62.30 & & 46.25 &  & 35.75 \\
      & \vline & 7B & \vline & Zoom Eye\,\textbf{(Ours)} & \vline & \ding{51} & \vline & \textbf{83.25} & & \textbf{53.25} &  & \textbf{50.75} \\
    \midrule
      \multirow{2}{*}{Qwen2.5-VL} & \vline & 7B & \vline & Pixel Reasoner & \vline & \ding{55} & \vline & 84.82 & & - &  & 66.00 \\
      & \vline & 3B & \vline & Zoom Eye\,\textbf{(Ours)} & \vline & \ding{51} & \vline & \textbf{89.01} & & - &  & \textbf{68.38} \\
    
    \bottomrule
    \end{tabular}
  \end{center}
  \vspace{-2mm}
  \caption{Performance comparison between Zoom Eye and DC$^2$~\cite{wang2024divide}, VisCrop~\cite{zhang2025mllms}, and Pixel Reasoner~\cite{pixelreasoner}.}
  \label{table:compare other methods}
  \vspace{-4mm}
\end{table*}

\begin{table}[t]
\footnotesize
  \begin{center}
  \resizebox{\hsize}{!}{
  \setlength\tabcolsep{5pt}
    \begin{tabular}{l c c c c c c c} 
    \toprule
      & \multirow{2}{*}{\makecell{Input\\Res.}} &\multirow{2}{*}{\makecell{Search\\ Res.}} &\multirow{2}{*}{\makecell{Zero\\shot}} & \multirow{2}{*}{\makecell{Indep.\\search}}&  &\multirow{2}{*}{\makecell{$V^*$\\Bench}} & \multirow{2}{*}{\makecell{HR\\Bench}}\\
      \textbf{Method} &    & &  & &  & &  \\
    \midrule
      $V^*$ search & 224 & 768 & \ding{55} & \ding{55} & \vline & 75.39 & 37.81  \\
      Zoom Eye & 224 & 224 & \ding{51} & \ding{51} & \vline & \textbf{81.58} & \textbf{47.63} \\
    \bottomrule
    \end{tabular}
    }
  \end{center}
  \caption{Performance comparison between Zoom Eye and $V^*$ Search~\cite{wu2024v}. Input Res.: The input resolution of the model generating the final response; Search Res.: The resolution required during the search process; Zero shot: Whether the method could be adapted for models without specialized additional training; Indep. search: Whether the method could be applied to an MLLM independently instead of requiring an additional search model.}
  \label{table:compare vstar}
  \vspace{-2mm}
\end{table}

\subsubsection{Zoom Eye vs. V$^*$}\label{sec:compare vstar}
$V^*$~\cite{wu2024v} is a LLM-guided search pipeline for MLLMs. To match the input resolution of the $V^*$ model, we specifically trained a 224px version of the LLaVA-v1.5 model for a fair comparison. Apart from using CLIP-224~\cite{radford2021learning} as the vision encoder, all other settings were identical to those of LLaVA-v1.5. 

From Table \ref{table:compare vstar}, it is evident that compared to $V^*$, our method offers several advantages: (1) The $V^*$ pipeline requires specifically targeted training data, making zero-shot searches impossible, whereas our method utilizes the native capabilities of MLLMs, allowing adaptation to any MLLM without additional training; (2) $V^*$’s search process necessitates the integration of another specially trained MLLM to guide the search, along with an extra high-resolution image encoder~\cite{minderer2022owlvit}(768px), while our approach operates at the native resolution of MLLMs and conducts searches independently; (3) Our method demonstrates superior performance.

\subsubsection{Zoom Eye vs. Others}
We also provide a comparison between Zoom Eye and DC$^2$~\cite{wang2024divide}, VisCrop~\cite{zhang2025mllms}, and Pixel Reasoner~\cite{pixelreasoner}. The results in Table~\ref{table:compare other methods} consistently demonstrate the superior performance of Zoom Eye. We provide a further discussion regarding the comparison between Zoom Eye and these methods in Appendix~\ref{sec:compare appendix}.

\subsection{Case Study}\label{sec:case}
We visualize some cases in Figure \ref{fig:case}, along with error examples mentioned in §\ref{sec:shortcomings}. We present cases for single type\,1 cue, multiple type\,1 cues, and type\,2 cue, which is corresponding to the examples in Table \ref{table:cues type}. From the figure, it can be observed that Zoom Eye accurately seeks out cues, enabling the MLLM to focus on the crucial visual information and respond to queries precisely.

\section{Related Work}
\textbf{Multimodal LLMs.} 
Since the advent of large language models\;(LLMs), they have achieved success across various linguistic applications, such as in-context learning~\cite{dong2022survey, zhang2022automatic, li2025taco, li2025m2iv} and retrieval augmented generation~\cite{liu2024much,zhao2024seer,zhao2024funnelrag}, which facilitated the emergence of Multimodal LLMs, with pioneering works including \cite{alayrac2022flamingo,li2023blip,koh2023grounding}. Following these, LLaVA~\cite{liu2024visual} employed GPT-4~\cite{achiam2023gpt} to develop training data, inspiring a series of works focused on visual instruction data~\cite{liu2024improved,instructblip,chen2023sharegpt4v}. Since these models utilize pretrained vision encoders~\cite{radford2021learning, zhai2023sigmoid} to process image, the resolution that MLLMs can handle is limited by the input resolution of these encoders. To address it, AnyRes was developed to flexibly manage varying resolutions~\cite{liu2024llavanext,chen2024internvl}. Additionally, there are efforts focused on utilizing high-resolution encoders~\cite{lu2024deepseek,wei2025vary} or investigating the selected layer of the encoders~\cite{chen2025rethinkingvisuallayerselection}. However, despite these efforts, the perception of the image by the MLLM remains as the original image itself. We hope to enable MLLMs to explore the varying hierarchical features of images to capture key information.

\noindent\textbf{Tree-based search.}
Tree-based search algorithms have been applied in text-only LLM reasoning and have demonstrated superior performance. Early works such as \cite{wei2022chain, wang2022self} relied on chain reasoning, a method susceptible to errors in one step propagating through subsequent steps. Consequently, ToT~\cite{yao2024tree} proposed a tree-based reasoning method that leverages the expansiveness of tree structures to widen the reasoning space.  Simultaneously, several similar studies were also introduced, which define a decomposed question step as a node and utilize beam search~\cite{xie2023decomposition} and Monte-Carlo Tree Search~\cite{hao2023reasoning} to uncover optimal solutions. Subsequently, TS-LLM~\cite{feng2023alphazero} utilized reinforcement learning to increase search depth, further enhancing reasoning performance. In our work, we conceptualize an image as a tree to search for crucial visual information using a specific algorithm. A close-related work is $V^*$~\cite{wu2024v}, and we describe the detailed comparison with it in §\ref{sec:compare vstar}.

\section{Limitations}

Although Zoom Eye offers several advantages, such as strong interpretability, model-agnostic, and training-free, it also comes with certain limitations. First, the current search procedure relies on heuristic strategies, including manually defined ranking functions and stopping criteria. While these designs are effective in many settings, they may not generalize optimally across all image types or task conditions. Second, the image is partitioned into fixed-size patches to construct the hierarchical tree structure, which may not align well with the semantic regions of the image. As a result, some visual cues may be fragmented or overlooked during traversal. Lastly, Zoom Eye is primarily tailored for natural images with spatially distributed visual elements. It is less applicable to document understanding tasks, where layout, reading order, and structured information (e.g., tables, forms) are central. Addressing these challenges---such as by integrating learnable search strategies or adaptive patch partitioning---will be an important direction for future work.

\section{Conclusion}
To address the limitations of text-level visual reasoning, we propose Zoom Eye, a type of vision-level reasoning method, a tree search algorithm designed to navigate the hierarchical and visual nature of images to capture detailed crucial information. Through prompts guiding MLLMs, we develop a ranking function and stopping criterion for Zoom Eye, which steers models to efficiently search along the image tree, seek out pertinent information, and accurately respond to related queries. Experiments show the broad-applicability and effectiveness of Zoom Eye, which substantially improves MLLMs' performance. Notably, Zoom Eye exhibits a test-time scaling phenomenon analogous to that observed in text-level reasoning. Meanwhile, through the analysis of failure cases, we identify several inherent limitations in current MLLMs’ visual reasoning capabilities, which we aim to address in future work.


\section{Acknowledgements}
This research is supported by National Key R\&D Program of China under grant (2022YFF0902600) and “Pioneer” and “Leading Goose” R\&D Program of Zhejiang (2023C01045).

\bibliography{acl_latex}

\appendix

\newpage
\tcbset{
  myboxstyle/.style={
    colback=gray!20,     
    colframe=black!70,    
    coltitle=white,       
    fonttitle=\bfseries,  
     fontupper=\itshape,
    boxrule=0.8mm,       
    arc=1mm,               
    boxsep=1mm,             
    left=1mm,              
    right=1mm,              
    top=1mm,               
    bottom=1mm,             
    toptitle=0mm,           
    bottomtitle=0mm,     
    enhanced,                  
  }
}

\section{Results of More MLLMs on High-Resolution Benchmark}
We present the results of additional MLLMs on high-resolution benchmarks in Table~\ref{table:appendix main results}, including models of smaller or larger scale. Consistent with the findings in the main paper, all evaluated models exhibit improved performance after being adapted to Zoom Eye, further demonstrating the effectiveness of vision-level reasoning in handling complex visual scenarios.

\begin{table*}[h!]
\scriptsize
  \begin{center}
    \begin{tabular}{l c c c c c c c c c c c} 
    \toprule
    & \multicolumn{3}{c}{$V^*$ Bench} & & \multicolumn{3}{c}{HR-Bench 4K} & & \multicolumn{3}{c}{HR-Bench 8K}\\
      \textbf{Model} & \textbf{Attr} & \textbf{Spatial} & \textbf{Overall} & & \textbf{FSP} & \textbf{FCP} & 
      \textbf{Overall} & & \textbf{FSP} & \textbf{FCP} & \textbf{Overall}\\
    \midrule
    \textbf{Baseline and Local Input Zoom Eye} & &  & & \vline &  &  &  & \vline &  &  & \\
      LLaVA-v1.5-13B~\cite{liu2024improved}         & 41.74 &  55.26 &  47.12& \vline &  45.25 &  41.25 &  43.25 & \vline &  37.50   &  38.0 &  37.75\\
      LLaVA-v1.5-13B w/ Zoom Eye & 87.83 & 81.58 & 85.34 & \vline & 73.0 & 43.25 &  58.13 & \vline &  67.25 &  45.50&  56.38 \\
      \rowcolor{gray!20}     
    \quad \quad \quad \quad $\Delta$ & \textcolor{darkgreen}{+46.09} & \textcolor{darkgreen}{+26.32} & \textcolor{darkgreen}{+38.22} & \vline & \textcolor{darkgreen}{+27.75} & \textcolor{darkgreen}{+2.00} & \textcolor{darkgreen}{+14.88} & \vline & \textcolor{darkgreen}{+29.75} & \textcolor{darkgreen}{+7.50} & \textcolor{darkgreen}{+18.63} \\
    \midrule
     \textbf{Baseline and Global+Local Input Zoom Eye} & &  & & \vline &  &  &  & \vline &  &  & \\
      
     LLaVA-ov-0.5B~\cite{li2024llava}            & 63.48 & 64.47 & 63.87 & \vline & 63.50 & 39.50 & 51.50 & \vline & 47.25 & 38.25 & 42.75\\
    
     LLaVA-ov-0.5B w/ Zoom Eye& 85.22 & 73.68 & 80.62 & \vline & 75.50 & 39.75 & 57.63 & \vline & 68.50 & 38.25 & 53.38\\
    \rowcolor{gray!20}     
    \quad \quad \quad \quad $\Delta$ & \textcolor{darkgreen}{+21.74} & \textcolor{darkgreen}{+9.21} & \textcolor{darkgreen}{+16.75} & \vline & \textcolor{darkgreen}{+12.00} & \textcolor{darkgreen}{+0.25} & \textcolor{darkgreen}{+6.13} & \vline & \textcolor{darkgreen}{+21.25} & +0.00 & \textcolor{darkgreen}{+10.63} \\
     \arrayrulecolor{gray}\hline  
     \arrayrulecolor{black}       

     InternVL2.5-4B~\cite{chen2024expanding}         & 69.57	&71.05	& 70.16 & \vline &  77.50&	53.75&	65.63 & \vline &  63.00	&49.25	& 56.13 \\
     InternVL2.5-4B w/ Zoom Eye & 85.22 &	77.63	& 82.20 & \vline & 81.25	& 56.75 &	69.00 & \vline &  80.00	&52.25&	66.13 \\
    \rowcolor{gray!20}     
    \quad \quad \quad \quad $\Delta$ & \textcolor{darkgreen}{+15.65} & \textcolor{darkgreen}{+6.58} & \textcolor{darkgreen}{+12.04} & \vline & \textcolor{darkgreen}{+3.75} & \textcolor{darkgreen}{+3.00} & \textcolor{darkgreen}{+3.37} &\vline & \textcolor{darkgreen}{+17.00} & \textcolor{darkgreen}{+3.00} & \textcolor{darkgreen}{+10.00} \\
    \arrayrulecolor{gray}\hline  
     \arrayrulecolor{black}       

     InternVL2.5-26B~\cite{chen2024expanding}         & 73.91	&72.37	& 73.30 & \vline &  82.00&	66.25&	74.13 & \vline &  73.00	&61.75	& 67.38 \\
     InternVL2.5-26B w/ Zoom Eye & 91.30 &	86.84	& 89.53 & \vline & 89.75	& 68.25 &	79.00 & \vline &  89.25	&63.00&	76.13 \\
    \rowcolor{gray!20}     
    \quad \quad \quad \quad $\Delta$ & \textcolor{darkgreen}{+17.39} & \textcolor{darkgreen}{+14.47} & \textcolor{darkgreen}{+16.23} &\vline & \textcolor{darkgreen}{+7.75} & \textcolor{darkgreen}{+2.00} & \textcolor{darkgreen}{+4.87} &\vline & \textcolor{darkgreen}{+16.25} & \textcolor{darkgreen}{+1.25} & \textcolor{darkgreen}{+8.75} \\
    \bottomrule
    \end{tabular}
  \end{center}
  \vspace{-3mm}
  \caption{Results of more models on high-resolution benchmarks.}
  \label{table:appendix main results}
  \vspace{-1mm}
\end{table*}

\section{Compared with Other HR Processing Methods}\label{sec:compare appendix}

\subsection{Zoom Eye vs. DC$^2$} 
DC$^2$~\cite{wang2024divide}\;(\textbf{D}ivide, \textbf{C}onquer, and \textbf{C}ombine) is a framework that supplements visual information  using text for high-resolution images understanding. Like our approach, it builds an image as a tree. The MLLM then generates textual descriptions for each leaf patch. These descriptions are then relayed to the parent nodes, which create combined descriptions by synthesizing the contents from their child nodes with their own. This process continues up to the root node.

Our approach differs from DC$^2$ in two key ways: (1)\;DC$^2$ uses textual modalities to supplement the missing visual information at high resolutions, whereas Zoom Eye employs simulated zooming operations, allowing the MLLM to actively discover missing visual details; (2)\;DC$^2$ is question-agnostic, generating descriptions consistently across different questions, which may lead to unfocused textual content. In contrast, Zoom Eye is question-driven in its visual cues searching, yielding more precise visual information that is instrumental in answering the input question. Table \ref{table:compare other methods} shows the better performance of Zoom Eye.

\subsection{Zoom Eye vs. Pixel Reasoner} 
Pixel Reasoner~\cite{pixelreasoner} is a multimodal model that combines curated reasoning trajectories with curiosity-driven reinforcement learning to enable effective zooming operations and significantly improve fine-grained visual reasoning.

The results on Table~\ref{table:compare other methods} demonstrate that:
(1) Zoom Eye outperforms Pixel Reasoner on both benchmarks, even with a smaller backbone (Qwen2.5VL-3B vs. Qwen2.5VL-7B), demonstrating its superior capability in enhancing vision-level visual reasoning within MLLMs;
(2) More importantly, Zoom Eye is entirely training-free, relying solely on prompting. In contrast, Pixel Reasoner requires constructing a supervised fine-tuning dataset pipeline and involves resource-intensive reinforcement learning.

This comparison underscores Zoom Eye’s core strength: achieving competitive or superior performance without any fine-tuning or task-specific training, making it a more adaptable solution in vision-level visual reasoning.

\subsection{Zoom Eye vs. VisCrop}

VisCrop crops and re-feeds the region focused by the attention map into the model -- essentially enabling the MLLM to “\textit{look again}” at a single focal point.

In contrast, Zoom Eye models the image as a tree, and guides the MLLM through a confidence-driven zoom-in process until a high-confidence answer node is found. This enables the MLLM to “\textit{look multiple times}” in a more structured and semantically informed way.

From Table~\ref{table:compare other methods}, we note that as resolution increases (from HR-4K to HR-8K), MKWTL’s performance degrades significantly, likely because a single “\textit{look again}” fails to capture fine-grained cues in these complex scenarios. In contrast, Zoom Eye maintains stable performance, showcasing the advantage of “\textit{look multiple times, until desirable cues are found to answer the question}”.

This comparison illustrates that MKWTL enables “\textit{a second glance}”, while Zoom Eye further enables “\textit{multi-step visual reasoning}”, being increasingly beneficial as visual complexity grows.

\section{Complete Results on MME-RealWorld Benchmark}\label{sec:appendix MME-RealWorld}

We provide the complete results of Zoom Eye on MME-RealWorld Benchmark~\cite{zhang2024mme}, as show in Table \ref{table:mme all}. This benchmark includes 5 data categories: Monitoring\;(MO), Autonomous Driving\;(AD), OCR, emote Sensing\;(RS), and Diagram and Table\;(TD). Since Zoom Eye is not applicable to the TD task,  we do not conduct tests on it. It could be observed that Zoom Eye improves the performance of LLaVA-ov-7B across most sub-tasks, with particularly significant improvements in certain tasks. For instance, it achieves a 20.22\% improvement in the Person$_\text{{color}}$ task, a 29.11\% improvement in the Motion$_\text{{vehicle}}$ task, and a 12.93\% improvement in the Visual$_\text{{trafficsignal}}$  task, demonstrating the effectiveness of Zoom Eye. However, performance declines were observed in some sub-tasks when using Zoom Eye. We have analyzed these cases in the main paper, revealing certain limitations of the employed MLLM. Addressing these issues will be a focus of our future work.

\begin{table}[h!]
\footnotesize
  \begin{center}
  \resizebox{\hsize}{!}{
  \setlength\tabcolsep{2pt}
    \begin{tabular}{l c c c c c r} 
    \toprule
      \textbf{Task} &  & & & LLaVA$_{\text{ov}}$-7B & +ZoomEye & \;$\Delta\uparrow$ \\
    \midrule
      \multirow{10}{*}{\textbf{MO}} & \vline & Calculate & \vline & 36.33 & 38.67 & \textcolor{darkgreen}{+2.34} \\
                  & \vline & Intention & \vline  & 27.55 & 38.78 & \textcolor{darkgreen}{+11.23}\\
                  & \vline & Property & \vline  & 55.0 & 60.0 & \textcolor{darkgreen}{+5.0}\\
                  & \vline & Vehicle$_{\text{counting}}$ & \vline  & 59.89 & 61.14 &\textcolor{darkgreen}{+1.25} \\
                  & \vline & Person$_{\text{counting}}$ & \vline  & 61.35 & 61.87 & \textcolor{darkgreen}{+0.52}\\
                  & \vline & Vehicle$_{\text{location}}$ & \vline  & 33.82 & 33.82 &  -\\
                  & \vline & Vehicle$_{\text{orientation}}$ & \vline  & 19.35 & 18.71 & \textcolor{red}{-0.64}\\
                  & \vline & Vehicle$_{\text{color}}$ & \vline  & 43.65 & 49.24 & \textcolor{darkgreen}{+5.59}\\
                  & \vline & Person$_{\text{color}}$ & \vline  & 24.72 & 44.94 & \textcolor{darkgreen}{+20.22}\\
                  & \vline & Person$_{\text{orientation}}$ & \vline  & 10.53 & 10.53 & - \\
    \midrule
      \multirow{15}{*}{\textbf{AD}} & \vline & Intention$_{\text{ego}}$ & \vline  & 28.62 & 28.95 & \textcolor{darkgreen}{+0.33} \\
                  & \vline & Intention$_{\text{pedestrian}}$ & \vline  & 52.43 & 53.40 & \textcolor{darkgreen}{+0.97} \\
                  & \vline & Intention$_{\text{vehicle}}$ & \vline  & 30.92 & 33.33 & \textcolor{darkgreen}{+2.41} \\
                  & \vline & Interaction$_{\text{other2other}}$ & \vline  & 12.94 & 13.43 & \textcolor{darkgreen}{+0.49} \\
                  & \vline & Attention$_{\text{trafficsignal}}$ & \vline  & 71.89 & 68.66 & \textcolor{red}{-3.23} \\
                  & \vline & Interaction$_{\text{ego2pedestrain}}$ & \vline  & 27.36 & 28.30 &\textcolor{darkgreen}{+0.94} \\
                  & \vline & Interaction$_{\text{ego2trafficsignal}}$ & \vline  & 22.86 & 25.71 & \textcolor{darkgreen}{+2.85} \\
                  & \vline & Interaction$_{\text{ego2vehicle}}$ & \vline  & 20.79 & 19.80 & \textcolor{red}{-0.99} \\
                  & \vline & Objects$_{\text{identify}}$ & \vline  & 64.40 & 64.85 & \textcolor{darkgreen}{+0.45} \\
                  & \vline & Motion$_{\text{vehicle}}$ & \vline  & 23.42 & 52.53 & \textcolor{darkgreen}{+29.11} \\
                  & \vline & Motion$_{\text{multivehicles}}$ & \vline  & 34.26 & 34.75 & \textcolor{darkgreen}{+0.49} \\
                  & \vline & Visual$_{\text{trafficsignal}}$ & \vline  & 60.20 & 73.13 & \textcolor{darkgreen}{+12.93} \\
                  & \vline & Motion$_{\text{pedestrain}}$ & \vline  & 34.15 & 40.85 & \textcolor{darkgreen}{+6.70} \\
                  & \vline & Object$_{\text{count}}$ & \vline  & 37.92 & 39.86 & \textcolor{darkgreen}{+1.94} \\
                  & \vline & Motion$_{\text{multipedestrians}}$ & \vline  & 31.24 & 31.64 & \textcolor{darkgreen}{+0.40} \\
    \midrule
     \multirow{7}{*}{\textbf{OCR}}             & \vline & Scene understanding & \vline  & 64.80 & 64.80 & - \\
                  & \vline & Character identification & \vline  & 57.60 & 56.40 & \textcolor{red}{-1.20}\\
                  & \vline & Adver \& product & \vline  & 76.64 & 78.37 & \textcolor{darkgreen}{+1.73} \\
                  & \vline & Book map poster & \vline  & 77.17 & 75.24 & \textcolor{red}{-1.93} \\
                  & \vline & License & \vline  & 80.16 & 82.39 & \textcolor{darkgreen}{+2.23} \\
                  & \vline &  Phone \& address & \vline  & 77.82 & 81.28 & \textcolor{darkgreen}{+3.46} \\
                  & \vline & Text recog & \vline  & 74.87 & 77.13 & \textcolor{darkgreen}{+2.26} \\
    \midrule
      \multirow{3}{*}{\textbf{RS}}          & \vline & Color & \vline  & 59.60 & 60.56 & \textcolor{darkgreen}{+0.96} \\
                & \vline & Count & \vline  & 32.95 & 35.56 & \textcolor{darkgreen}{+2.61}\\
                & \vline &  Position & \vline  & 61.40 & 48.45 & \textcolor{red}{-12.95} \\
    \bottomrule
    \end{tabular}
    }
  \end{center}
  \caption{Performance comparison between Zoom Eye and the baseline model on MME-RealWorld benchmark. MO (Monitoring), AD (Autonomous Driving), OCR and RS (Remote Sensing) are data categories within this benchmark.}
  \label{table:mme all}
\end{table}

\section{Implementation Details}\label{sec:appendix Implementation}
Due to the page limit of the main paper, we provide more implementation details here. In §\ref{sec:appendix local} and §\ref{sec:appendix global+local},  we detail the implementation of Local Input and Global+Local Input, respectively. §\ref{sec:additional} describes the implementations common to both. Finally, based on the introductions in the first three subsections, we present the complete algorithm workflow of Zoom Eye in §\ref{sec:complete}.

\subsection{Local Input}\label{sec:appendix local}
We select LLaVA-v1.5-7B~\cite{liu2024improved} and 13B as our MLLMs, with the vision encoder's input resolution as 336px and naive processing. We set the threshold of the stopping criterion at $\tau$ = 0.8 and define the weighted function as $\mathcal{W}$ = $ \frac{1-b}{D^2}\times d^2 + b $, where $D$ denotes the depth of the image tree, $d$ is the depth of the visited node during the search, and $b$ is a bias value, set here at 0.2. The prompt templates for calculating \textit{existing confidence}, \textit{latent confidence}, and \textit{answering confidence}~(please refer to §\ref{sec:confidence} and §\ref{sec:stop} for the discussion on these three confidence values) are set as:

\begin{tcolorbox}[
    myboxstyle, 
    title=Prompt Templates of Local Input  
]
\begin{itemize}
\setlength{\itemsep}{4pt}
  \item $\text{p}_e(o)$: \textless local patch\textgreater\,Is there a $\{o\}$ in the image? Answer Yes or No.
  \item $\text{p}_l(o)$: \textless local patch\textgreater\,According to your common sense knowledge and the content of the image, is it possible to find a $\{o\}$ in the image? Answer Yes or No and tell the reason.
  \item $\text{p}_a(q)$: \textless local patch\textgreater\,Question: $\{q\}$ \textbackslash nCould you answer the question based on the the available visual information? Answer Yes or No.
\end{itemize}
\end{tcolorbox}

\noindent where the $o$ and $q$ are the input visual cue and question, which could be referred to §\ref{sec:algorithm} .

\begin{figure}[ht!]
    \centering
    \includegraphics[width=8.2cm]{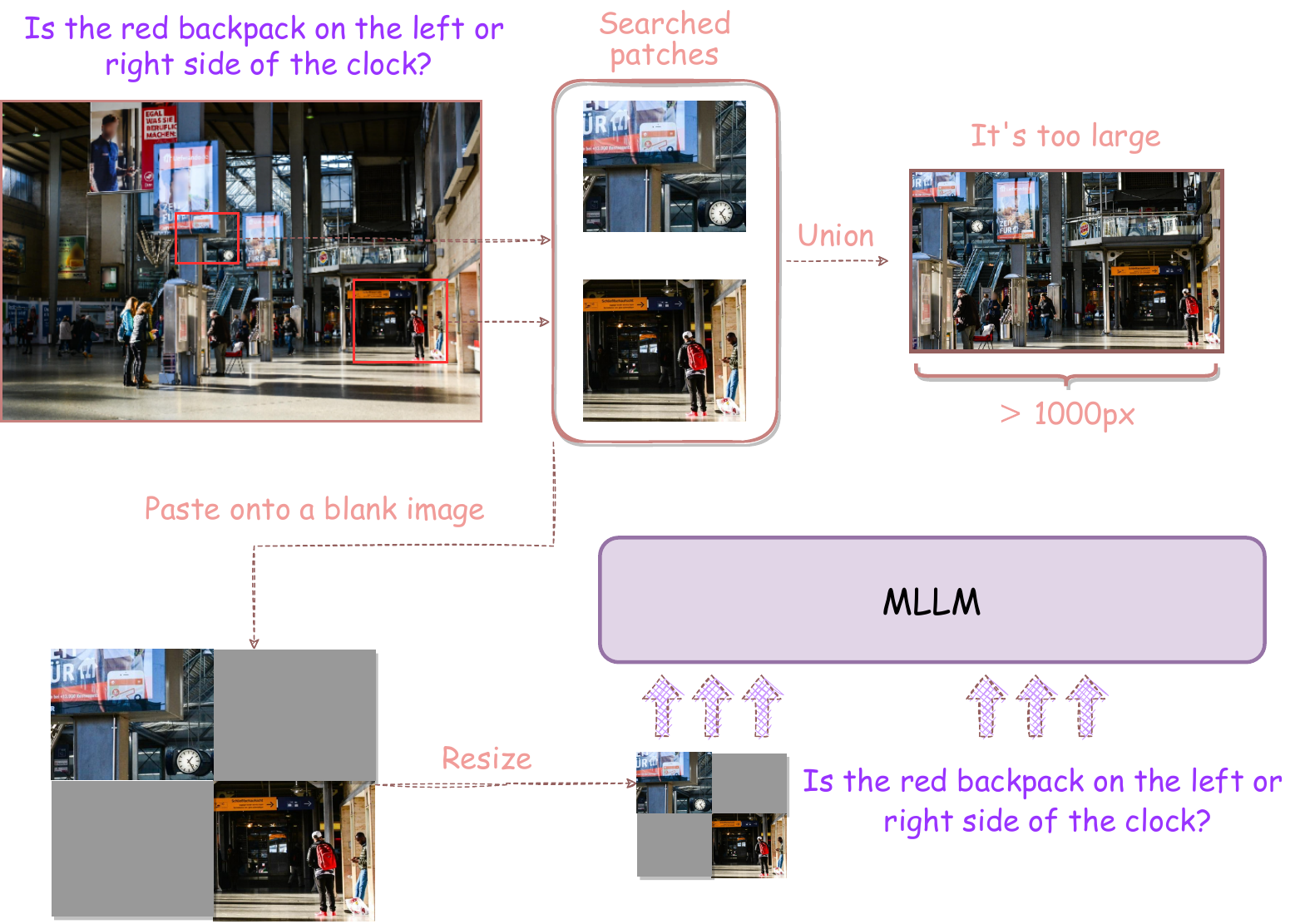}
    \caption{If the area of the union bounding box is too large, we paste the searched patches onto a blank image according to their relative positions in the original image, and then feed it to the MLLMs.\textbf{It is notable that this operation is only applied to Local Input, while for Local+Global, we consistently provide the MLLMs with the full union patch as input.}}
    \label{fig:patch paste}
    \vspace{-3mm}
\end{figure}

As mentioned in §\ref{sec:algorithm}, the final visual input uses the union of all searched patches. However, when multiple distant patches are combined, they may form a large image. For MLLMs using naive resize processing, information can still be lost during downsampling. Therefore, for the Local Input Zoom Eye with naive resize processing, when the area of  $b^*$  is relatively large (with the longer side exceeding 1000px), we skip the Union operation.  Instead, we paste the searched patches onto a blank image according to their relative positions in the original image, and then feed it to the MLLMs. An example is shown in Figure \ref{fig:patch paste}.

\subsection{Global\,+\,Local Input}\label{sec:appendix global+local}
We select LLaVA-ov(oneVision)-0.5B~\cite{li2024llava} and 7B as our MLLMs, with the vision encoder's input resolution as 384px and AnyRes processing. We define the maximum AnyRes block as 12, set $\tau$ at 0.6 and define $\mathcal{W}$ as $ \frac{1-b}{D^2}\times d^2 + b $, where $D$ denotes the depth of the image tree, $d$ is the depth of the visited node during the search, and $b$ is a bias value, set here at 0.6. The prompt templates for calculating \textit{existing confidence}, \textit{latent confidence}, and \textit{answering confidence} are set as:

\begin{tcolorbox}[
    myboxstyle, 
    title=Prompt Templates of Global\,+\,Local Input
]
\begin{itemize}
\setlength{\itemsep}{4pt}
  \item $\text{p}_e(o)$: \textless global image\textgreater \textless local patch\textgreater\,Is there a $\{o\}$ in the zoomed-in view? Answer Yes or No.
  \item $\text{p}_l(o)$: \textless global image\textgreater \textless local patch\textgreater\,According to your common sense knowledge and the content of the zoomed-in view, along with its location in the image, is it possible to find a $\{o\}$ by further zooming in the current view? Answer Yes or No and tell the reason.
  \item $\text{p}_a(q)$: \textless global image\textgreater \textless local patch\textgreater\,Question: $\{q\}$ \textbackslash nCould you answer the question based on the the available visual information? Answer Yes or No.
\end{itemize}
\end{tcolorbox}

\subsection{Additional Settings}\label{sec:additional}
For both input implementation, we set the maximum search depth at 2 when searching for type\,2 cues to save costs. In §\ref{sec:algorithm}, we state that we search the whole tree to add all nodes with sufficient \textit{existing confidence} to $L$ if  type\,2 cue is generated. Thus, we introduce an additional threshold $\tau_2$ for this condition, which is set at 0.8 for both implementation. The decomposed question template $\text{p}_{dq}(o_i)$ is assigned as ``\texttt{What is the appearance of the $\{o_i\}$?}". For type\,1 search, a key aspect is determining the value of $\tau$. If it is set too low, an incorrect patch, which probably lead to erroneous guidance  for MLLMs, may be selected. Conversely, setting $\tau$ too high, surpassing the $c_a$ values of all nodes in the tree, would compel MLLMs to search the entire tree unnecessarily, thus wasting time. Therefore, we adopt a strategy where $\tau$ is progressively reduced as the number of search steps increases. Specifically, if the number of  search steps exceeds the step threshold $C$, we reduce the value of $\tau$ by 0.1. This reduction  occurs every $\delta$ steps, until the $c_a$ value of a node having been visited surpasses $\tau$ or $\tau$ falls below a predefined minimum limit $\tau_{min}$. For both implementation, we set $\delta$ at 2, $\tau_{min}$ at 0, and $C$ as $D\times3$. Finally, the in-context examples we utilized to generate visual cues are denote as $(q^{(1)},\textbf{o}^{(1)},\dots,q^{(m)},\textbf{o}^{(m)})$ and are presented at the end of this document.

\subsection{Complete Algorithm Workflow}\label{sec:complete}

With the aforementioned notation and description in place, we provide the complete algorithm workflow in Algorithm \ref{algorithm:complete}, where the Zoom Eye search method is shown in Algorithm \ref{algorithm:zoom eye}.

\begin{algorithm}
\caption{Complete Algorithm Workflow of Zoom Eye}
\small
\begin{algorithmic}[1]
\Require Multimodal LLM $\Phi_{\theta}$, input question-image pair (\textbf{I}, $q$), decomposed question template $\text{p}_{dq}$, in-context examples $(q^{(1)},\textbf{o}^{(1)},\dots,q^{(m)},\textbf{o}^{(m)},q)$
\State $\{o_1,\dots,o_k\} \gets \Phi_{\theta}.\text{generate}(q^{(1)},\textbf{o}^{(1)},\dots,q^{(m)},\textbf{o}^{(m)},q)$
\State  Initialize $L$ as the empty list
\State  Build \textbf{I} as a tree $T$
\For{$i=1,\dots,k$}
    \If{$k==1$}
        \State $q_s \gets q$
    \Else
        \State $q_s \gets \text{p}_{dq}(o_i)$
    \EndIf
    \State $L$.extend(\Call{Zoom Eye}{$T, o_i, q_s$})
\EndFor
\State $\textbf{b}^* \gets \text{Union bounding-boxes of all nodes in } L$
\State $n^* \gets \{\textbf{I}, \textbf{b}^*\}$
\State Final response $\gets \Phi_{\theta}.\text{generate}(\mathcal{V}(n^*),q)$
\end{algorithmic}\label{algorithm:complete}
\end{algorithm}

\begin{algorithm}
\caption{Zoom Eye Search}
\small
\begin{algorithmic}[1] 
\Require{Threshold of type 1 cue and type 2 cue ($\tau, \tau_2$), minimum limit $\tau_{min}$, interval $\delta $}
\Function{Zoom Eye}{\;$T,o_i,q_s$}
    \State Initialize $Q$ as the empty queue \{\}
    \State $Q$.\texttt{append($T$.root)}
    \State Initialize $L_i$ as the empty list
    \State search all $\gets$ $o_i$.\texttt{startswith}(``all")
    \If{not search all}
        \State \Call{Zoom Eye Type 1}{\;$T,Q,L_i,q_s,\tau$}
    \Else
        \State \Call{Zoom Eye Type 2}{\;$Q,L_i,\tau_2$}
    \EndIf
    \State \Return{$L_i$}
\EndFunction
\State
\Function{Zoom Eye Type 1}{\;$T,Q,L_i,q_s,\tau$}
    \State import $\mathcal{R}$ and $\mathcal{S}$ from Algorithm \ref{algorithm:rank}
    \State count $\gets$ 0
    \State $C \gets T.depth \times 3$
    \State Initialize $n_m$ as $T$.root to record the node with the best $c_a$
    \While{$Q$ is not empty}
        \State $n_t \gets Q.\texttt{pop()}$
        \State $N$.\texttt{append}($n_t$)
        \State count $\gets$ count + 1
        \If{count $\geq$ C}
            \State $\tau \gets \tau - 0.1$
            \State $C \gets C+\delta$
            \If{$\tau < \tau_{min}$}
                \State \textbf{break}
            \EndIf
        \EndIf
        \If{$\mathcal{S}(n_t,q_s,\tau)==\text{True}$}
            \State $L_i$.\texttt{append}($n_t$)
            \State \textbf{break}
        \ElsIf{$\mathcal{S}(n_m,q_s,\tau)==\text{True}$}
            \State $L_i$.\texttt{append}($n_m$)
            \State \textbf{break}
        \EndIf
        \If{$n_t.c_a \geq n_m.c_a$}
            \State $n_m \gets n_t$
        \EndIf
        \State $s \gets n_t.\texttt{children.size}$
        \For{$j=1,\dots,s$}
            \State $Q$.\texttt{append($n_t$.children[j])}
        \EndFor
        \State $Q$.\texttt{sort($\mathcal{R}(o_i)$)}
    \EndWhile
\EndFunction
\State
\Function{Zoom Eye Type 2}{\;$Q,L_i,\tau_2$}
    \While{$Q$ is not empty}
        \If{$n_t.\texttt{depth} \geq 2$}
            \State \textbf{break}
        \EndIf
        \State $c_e$ $\gets$ calculate the existing confidence of $n_t$
        \If{$c_e \geq \tau_2$}
            \State $L_i$.\texttt{append}($n_t$)
        \EndIf
        \State $s \gets n_t.\texttt{children.size}$
        \For{$j=1,\dots,s$}
            \State $Q$.\texttt{append($n_t$.children[j])}
        \EndFor
    \EndWhile
\EndFunction
\end{algorithmic}\label{algorithm:zoom eye}
\end{algorithm}

\begin{tcolorbox}[myboxstyle, 
title=In-Context Examples of $V^*$ Bench,
before upper={\small\baselineskip=10pt}]
\begin{itemize}[label={}]
\setlength{\itemsep}{6pt}
  \item \textbf{User:} Question: What is the color of the boy's bag? If you want to answer the question, which objects' information do you need?
  \item \textbf{Assistant:} To answer the question, I need know the location of the boy with a bag so that I can determine the color of the bag. So I need the information about the following objects: boy with a bag.
  \item \textbf{User:} Is the yellow car on the left or right side of the white car?
  \item \textbf{Assistant:} To answer the question, I need know the location of the yellow car and the white car so that I can determine the positional relationship between the two of them. So I need the information about the following objects: white car and yellow car.
  \item \textbf{User:} Tell me the number on the black board.
  \item \textbf{Assistant:} To answer the question, I need know the location of the black board so that I can determine the number on it. So I need the information about the following objects: black board
  \item \textbf{User:} Is the girl with pink hair on the left or right side of the man with backpack?
  \item \textbf{Assistant:} To answer the question, I need know the location of the girl with pink hair and the man with backpack so that I can determine the positional relationship between the two of them. So I need the information about the following objects: girl with pink hair and man with backpack.
  \item \textbf{User:} What kind of animal is on the red sign?
  \item \textbf{Assistant:} To answer the question, I need know the location of the red sign so that I can determine the kind of animal on it. So I need the information about the following objects: red sign.
  \item \textbf{User:} From the information on that advertising board, what is the type of this shop?
  \item \textbf{Assistant:} To answer the question, I need know the location of the advertising board so that I can determine the type of the shop. So I need the information about the following objects: advertising board.
\end{itemize}
\end{tcolorbox}

\begin{tcolorbox}[myboxstyle, 
title=In-Context Examples of HR-Bench,
before upper={\small\baselineskip=10pt}]
\begin{itemize}[label={}]
\setlength{\itemsep}{4pt}
  \item \textbf{User:} Question: What is the color of the boy's bag? If you want to answer the question, which objects' information do you need?
  \item \textbf{Assistant:} To answer the question, I need know the location of the boy with a bag so that I can determine the color of the bag. So I need the information about the following objects: boy with a bag.
  \item \textbf{User:} Is the yellow car on the left or right side of the white car?
  \item \textbf{Assistant:} To answer the question, I need know the location of the yellow car and the white car so that I can determine the positional relationship between the two of them. So I need the information about the following objects: white car and yellow car.
  \item \textbf{User:} Tell me the number on the black board above the dog.
  \item \textbf{Assistant:} To answer the question, I need know the location of the black board above the dog so that I can determine the number on it. So I need the information about the following objects: black board above the dog.
  \item \textbf{User:} Is the girl with pink hair on the left or right side of the man with backpack?
  \item \textbf{Assistant:} To answer the question, I need know the location of the girl with pink hair and the man with backpack so that I can determine the positional relationship between the two of them. So I need the information about the following objects: girl with pink hair and man with backpack.
  \item \textbf{User:} What kind of animal is on the red sign?
  \item \textbf{Assistant:} To answer the question, I need know the location of the red sign so that I can determine the kind of animal on it. So I need the information about the following objects: red sign.
  \item \textbf{User:} How many cars in the image?
  \item \textbf{Assistant:} To answer the question, I need know the location of all cars so that I can determine the number of cars. So I need the information about the following objects: all cars.
  
\end{itemize}
\end{tcolorbox}


\end{document}